\documentclass[lettersize,journal]{IEEEtran}
\usepackage{amsmath,amsfonts}
\usepackage{algorithmic}
\usepackage{algorithm}
\DeclareUnicodeCharacter{2265}{\ensuremath{\geq}}
\usepackage{array}
\usepackage{tabularx}
\usepackage{colortbl}  
\usepackage{xcolor}
\usepackage{hyperref}
\hypersetup{
    colorlinks=true,
    linkcolor=blue,
    filecolor=magenta,      
    urlcolor=cyan,
    pdftitle={Overleaf Example},
    pdfpagemode=UseOutlines,
    }

\usepackage{textcomp}
\usepackage{stfloats}
\usepackage{url}
\usepackage{verbatim}
\usepackage{forest}

\usepackage{ifpdf}
\usepackage{graphicx}
\usepackage{caption}
\usepackage{subcaption}
\usepackage{threeparttable}
\usepackage{tikz}
\usetikzlibrary{positioning, shapes.geometric, arrows.meta}

\ifpdf
  \graphicspath{ {./fig/} }
  \DeclareGraphicsExtensions{.pdf,.jpeg,.png}
\else
  \graphicspath{ {./fig/} }
  \DeclareGraphicsExtensions{.eps}
\fi

\usepackage{cite}
\begin{document}

\title{A Survey on Radar-Based Fall Detection}

\author{Shuting Hu,
        Siyang Cao,
        Nima Toosizadeh,
        Jennifer Barton,
        Melvin G. Hector,
        Mindy J. Fain
\thanks{S. Hu and S. Cao are with the Department of Electrical and Computer Engineering, The University of Arizona, Tucson, AZ, 85721 USA. e-mail: (shutinghu@arizona.edu, caos@arizona.edu).}
\thanks{N. Toosizadeh is with the Department of Rehabilitation and Movement Sciences, Rutgers School of Health, Rutgers University. e-mail: (nima.toosizadeh@rutgers.edu).}
\thanks{J. Barton is with the Department of Biomedical Engineering, The University of Arizona, Tucson, AZ, 85721 USA. e-mail: (barton@arizona.edu).}
\thanks{M.G. Hector and M.J. Fain are with the Department of Medicine, The University of Arizona, Tucson, AZ, 85724 USA. e-mail: (Melvin.hector@banner.health.com, MFain@aging.arizona.edu).}
\thanks{This work was supported by National Institute of Biomedical Imaging and Bioengineering (NIBIB), National Institute of Health (NIH), Grant No. R21EB033454.}
}

\markboth{}%
{Hu \MakeLowercase{\textit{et al.}}: A Survey on Radar-Based Fall Detection}


\maketitle

\begin{abstract}
Fall detection, particularly critical for high-risk demographics like the elderly, is a key public health concern where timely detection can greatly minimize harm. With the advancements in radio frequency technology, radar has emerged as a powerful tool for human detection and tracking. Traditional machine learning algorithms, such as Support Vector Machines (SVM) and k-Nearest Neighbors (kNN), have shown promising outcomes. However, deep learning approaches, notably Convolutional Neural Networks (CNN) and Recurrent Neural Networks (RNN), have outperformed in learning intricate features and managing large, unstructured datasets.
This survey offers an in-depth analysis of radar-based fall detection, with emphasis on Micro-Doppler, Range-Doppler, and Range-Doppler-Angles techniques. We discuss the intricacies and challenges in fall detection and emphasize the necessity for a clear definition of falls and appropriate detection criteria, informed by diverse influencing factors.
We present an overview of radar signal processing principles and the underlying technology of radar-based fall detection, providing an accessible insight into machine learning and deep learning algorithms. After examining 74 research articles on radar-based fall detection published since 2000, we aim to bridge current research gaps and underscore the potential future research strategies, emphasizing the real-world applications possibility and the unexplored potential of deep learning in improving radar-based fall detection.
\end{abstract}


\section{INTRODUCTION}

\IEEEPARstart{T}{he} global rise in life expectancy, as reported by the World Health Organization (WHO)~\cite{WHO2021Ageing}, has led to an aging population and increased risk of falls among the elderly. The U.S. Centers for Disease Control and Prevention (CDC)~\cite{CDC2020Falls} estimates that annually, one in four older adults in the United States experiences a fall. These falls, as the leading cause of injury among older adults~\cite{bergen2016falls}~\cite{moreland2020trends}, contribute significantly to unintentional injuries and deaths. 

Three decades ago, the National Institutes of Health (NIH)~\cite{berg1992falls} identified fractures as the most common serious injuries from falls in older persons, making them more susceptible to falls. The fear of falling can also reduce the confidence of older adults in outdoor activities and their ability to live independently~\cite{schoene2019systematic}. The CDC reports the annual medical costs related to non-fatal fall injuries is about \$50 billion and \$754 million is spent related to fatal falls~\cite{CDC2020CostofFalls}. The growing elderly population and subsequent increase in falls could strain healthcare systems due to resource scarcity. The need for efficient fall detection systems is evident, as prompt detection and reporting of falls can significantly reduce associated risks.

Over the years, various methodologies have been devised to address this concern, including (i) Sound-based approaches~\cite{popescu2008acoustic}~\cite{li2012microphone}~\cite{khan2015unsupervised}~\cite{adnan2018fall}~\cite{lian2021fall} make use of acoustic signals or noises generated during a fall event. Microphones or wearable sound sensors detect these unique sound patterns, differentiating between regular noises and those produced by falls. (ii) Motion-based methods~\cite{chen2006wearable}~\cite{pierleoni2015high}~\cite{jung2015wearable}~\cite{mekruksavanich2022wearable} primarily use wearable devices equipped with accelerometers or gyroscopes. These sensors measure changes in velocities and orientations, identifying patterns consistent with falls. (iii) Vision-based systems~\cite{auvinet2010fall}~\cite{stone2014fall}~\cite{bian2014fall}~\cite{chen2020fall}~\cite{ramirez2021fall} utilize cameras to monitor individuals continuously. Through image processing and machine learning, these systems can discern between regular movements and falls.

Non-radar sensors, including sound, motion, and vision-based solutions, offer innovative ways to detect falls. Sound sensors are ubiquitous and non-intrusive but can generate false alarms due to ambient noise~\cite{popescu2008acoustic} and may raise privacy concerns. Motion sensors are compact and respect privacy, but they require frequent charging~\cite{kim2021self} and may cause discomfort~\cite{shimura2022engineering}. Vision sensors are non-contact, and benefit from advanced computer vision techniques, but they may intrude on privacy~\cite{mastorakis2014fall} and could fail to detect falls in case of occlusions~\cite{auvinet2010fall}. Despite their advantages, each sensor type has inherent limitations, often making hybrid systems or combined sensors an optimal choice for effective fall detection~\cite{liang2021collaborative}. 

Prior surveys on fall detection have been explored. Studies~\cite{noury2007fall}~\cite{ramachandran2020survey}~\cite{stavropoulos2020iot} focused on motion-based fall detection using sensors. Meanwhile, authors~\cite{mubashir2013survey}~\cite{igual2013challenges}\cite{biswas2020literature}~\cite{islam2020deep}~\cite{singh2022systematic} observed the rise of vision-based detectors and machine learning while noting performance and privacy issues. Zhang Zhong, et al~\cite{zhang2015survey} discussed the limitations of vision-based techniques and the benefits of depth sensors. Xu Tao, et al.~\cite{xu2018new} identified a transition towards Radio Frequency (RF) sensor-based methods. Recent reviews~\cite{wang2020elderly}~\cite{singh2020sensor}~\cite{nooruddin2022sensor}~\cite{tanwar2022pathway} underscored the need for real-time detection, real fall datasets, and privacy preservation, noting the rising use of radar for fall detection. These articles provide vital insights into the state and challenges of current fall detection technologies.

RF sensor-based methods for fall detection primarily encompass Wi-Fi and radar technologies. While Wi-Fi has achieved considerable milestones in this domain ~\cite{palipana2018falldefi}~\cite{zhao2018through}~\cite{ding2020wifi}~\cite{nakamura2020wi}~\cite{yang2022rethinking}~\cite{nakamura2022wi}, it falls beyond the scope of this paper. Radar carries great potential to be one of the leading technologies in the near future~\cite{amin2016radar}~\cite{amin2017radar}. They are non-invasive and ensure privacy since they don't capture visual images~\cite{sengupta2020mm}. They can detect through obstructions like walls~\cite{li2012through}, are unaffected by environmental conditions such as lighting~\cite{sengupta2020mm}, and can operate efficiently in low power~\cite{vandersmissen2018indoor}~\cite{almalioglu2020milli}. Additionally, they provide broad coverage while ensuring continuous monitoring~\cite{zhang2018real}~\cite{jin2019multiple}. 

\begin{figure}[htbp]
\centerline{\includegraphics[width=0.5\textwidth]{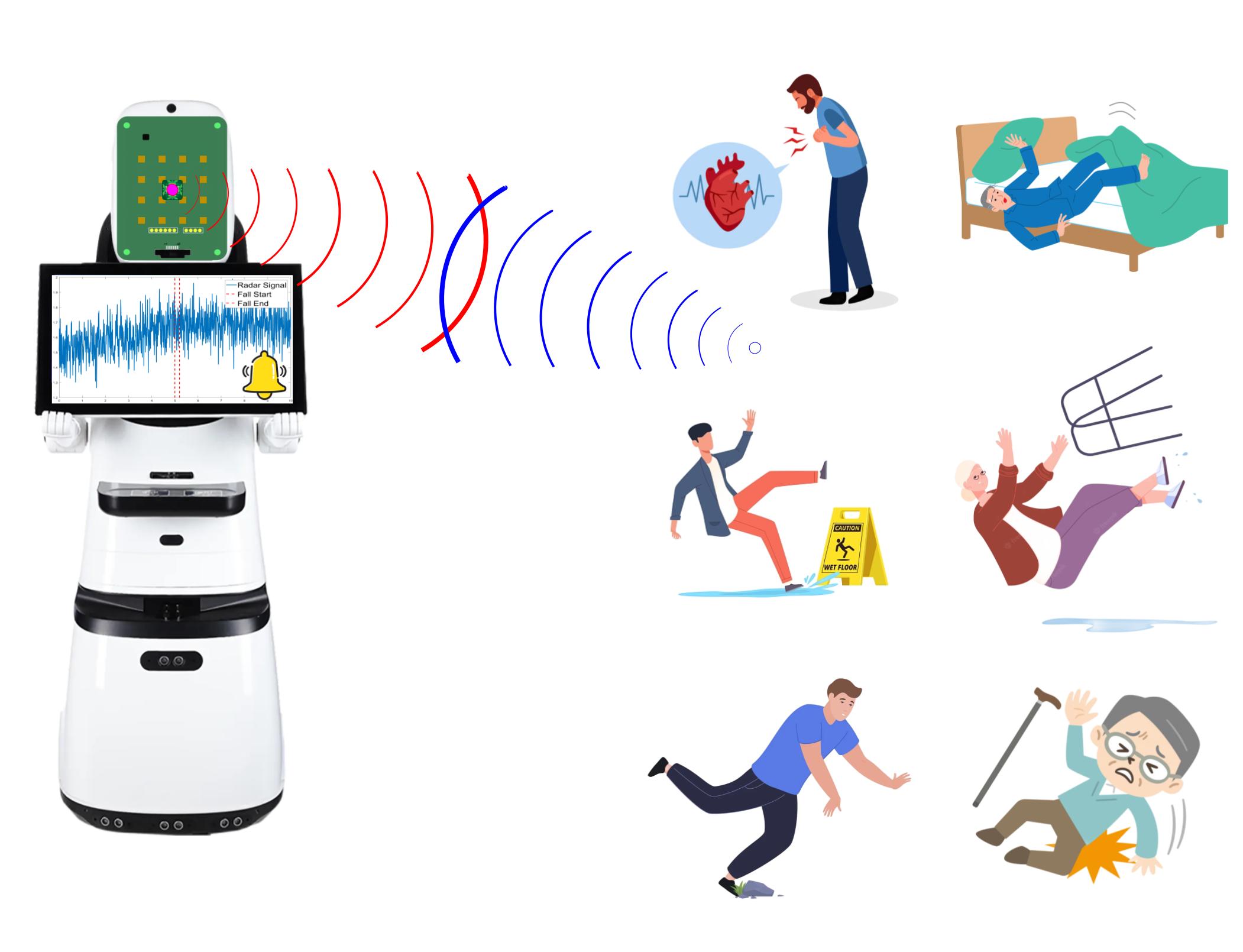}}
\caption{A radar-based fall detection robotic system. The radar device emits waves that interact with the environment. As a person moves or falls, the returning wave patterns change. These changes are captured and processed by the radar device, and sent to a computer or screen. The computer processes and visualizes the signal data, highlighting the point of a detected fall and trigger an alarm.}
\label{concept}
\end{figure}

Despite significant advancements in radar sensor technology, no prior surveys specifically delve into the evolution and potential of radar-based fall detection. While the developments in radar-based fall detection over the past two decades are vast, there remains a void in consolidating this information into a comprehensive survey. This paper aims to fill that gap. To provide a clearer understanding, Figure~\ref{concept} depicts a conceptual representation of a radar-based fall detection robotic system, showcasing how changes in a person's movements, such as falls, can be detected and processed using radar technology.

When selecting papers for our review, we established specific criteria to ensure the relevance and quality of the studies chosen: 
(i) Our primary focus was on papers that directly tackled themes related to ``radar fall" in conjunction with terms such as ``detection" and ``detector". 
(ii) We concentrated on publications from the past two decades. 
(iii) Emphasis was placed on studies that showcased clear methodologies and thorough data analyses. We excluded articles that primarily delved into hardware or antenna design. 
(iv) Furthermore, to recognize influential work, we filtered out articles with fewer than 10 citations before 2010 and those with under 5 citations prior to 2015. 

To the best of our knowledge, this is the first survey paper specifically focused on the two-decade evolution of radar-based fall detection. The rest of the paper is organized as follows: Section II provides an understanding of falls. Section III discusses radar detection system fundamentals with associated techniques. Section IV reviews radar-based fall detection in Micro-Doppler, Range-Doppler, and Range-Doppler-Angle approaches. Section V discusses existing limitations and suggests directions for future research. Section VI concludes the paper.

\section{Understanding Falls}

\subsection{The Challenges of Fall Detection}
Detecting a fall precisely is challenging due to its unpredictable and sudden nature. A fall can be defined as unintentionally coming to the ground or a lower level without a violent blow, loss of consciousness, or sudden paralysis~\cite{mj1987prevention}~\cite{zecevic2006defining}~\cite{noury2007fall}.

\subsubsection{Factors and Variables Related to Falls}
Extensive research has been conducted into the factors and variables causing falls, revealing numerous demographic, psychological, functional, medical, and physical characteristics associated with falls. Intrinsic risk factors relate to an individual's functional and health status including muscle strength, balance, motion, and sleep disorders~\cite{brassington2000sleep}~\cite{kvelde2013depressive}~\cite{tinetti1988risk}~\cite{anderson2020characteristics}. Extrinsic risk factors include environmental hazards such as slippery floors, poor lighting, improper use of assistive devices, and inappropriate footwear~\cite{boelens2013risk}. Most researchers conclude that falls and recurrent falls result from a combination of
intrinsic and extrinsic factors. The interrelation of these variables and the diversity of causes pose significant challenges to addressing the problem of falls. 

\subsubsection{Challenges of Public Datasets}
Capturing the moment of a fall is difficult due to the non-periodic nature of falls, leading to a scarcity of public datasets for fall incidents, especially real-life datasets. A few studies ventured into collecting and analyzing real fall events~\cite{vlaeyen2013fall}~\cite{robinovitch2013video}~\cite{rapp2012epidemiology}, while others choose to simulate the behavior of an elderly person~\cite{kwolek2014human}~\cite{h2qp48-16}~\cite{alzahrani2017fallfree}. Notable datasets in the radar domain, such as studies by Su, Bo Yu, et al.~\cite{su2014doppler}, Li, Haobo, et al. ~\cite{li2020distributed}, Yao, Yicheng, et al. ~\cite{yao2022fall}, Wang, Bo, et al. ~\cite{wang2022millimeter}, revealed temporal and scene diversity in falls, which complicates the process of data collection and annotation.

\subsection{Fall Categories and Scenarios}
Falls can occur in various settings, including living spaces, sleeping quarters, hallways, bathrooms, etc. Activities preceding a fall can range from transitioning on/off the bed, sofa, or wheelchair, walking, bending, dressing, bathing, to sitting and standing. Capturing genuine fall incidents is a lengthy process that's both time-consuming and labor-intensive. Given the unpredictable nature of falls, especially among the elderly, gathering a sufficient amount of authentic fall data for research requires extended periods of observation. This prolonged data collection process not only demands significant time but also incurs high human resource costs. Moreover, the ethical implications surrounding the collection of real fall data from vulnerable individuals, particularly the elderly, further complicates the process. It is this combination of ethical concerns and the logistical challenges of long-term data collection that underscores the reliance on simulated fall data in most existing research. Hence, most of the existing radar-based fall detection research relies on simulated actions with data collection based on the researchers' definitions. Upon reviewing established radar-based fall detection research, a recurring issue becomes evident: the absence of a standardized measure.

With the inherent challenges in designing and covering all potential real-life scenarios due to the limited range of activities, there's a pressing call for researchers to aim for realism in the collected data, ensuring it accurately mirrors actual fall situations. A possible approach includes setting up lab environments that closely mimic real-life settings with similar room sizes, furniture, and other elements. We recommend including diverse scenarios - dining room, living room, bedroom, and restroom - with specific activities defined within each. Participants should be encouraged to perform their daily movements naturally, and the age distribution of participants should be broad and not limited to younger individuals. Lastly, including individuals who require assistive devices like canes, walkers, or wheelchairs can enhance the realism and applicability of the study.

\subsection{Quality Metrics for Fall Detection} \label{Quality Metrics}
To develop a fall detection system applicable to real-life scenarios, it is crucial to minimize the false alarm. Evaluation metrics for this system can be derived from the realm of binary classification statistics, which includes accuracy, precision, recall, and the F1 score. In this context, ``items of interest" correspond to falls, with ``positive" indicating fall events and ``negative" non-fall events. A correct prediction is labeled ``true", whereas an incorrect prediction is ``false".

To simplify, we can define:

\begin{enumerate}
    \item True Positive (TP): A fall event is correctly identified as a fall by the device.
    \item False Positive (FP): A non-fall event is incorrectly identified as a fall by the device.
    \item True Negative (TN): A non-fall event is correctly identified as non-fall by the device.
    \item False Negative (FN): A fall event is incorrectly identified as non-fall by the device.
\end{enumerate}
Thus, items predicted correctly include TP and TN; items of interest predicted encompass TP and FP; and items of interest comprise TP and FN.

Given the likely imbalance between Activities of Daily Living (ADL) and fall samples in fall detection tasks, accuracy ($Acc=\frac{TP+TN}{TP+FP+TN+FN}$) may not be the most suitable performance measure. Precision ($Prec=\frac{TP}{TP+FP}$) and recall ($Rec=\frac{TP}{TP+FN}$), however, offer more meaningful evaluation metrics. 

Precision quantifies the proportion of identified falls that are actual falls, emphasizing the reduction of false alarms. A score of 100.0\% in precision signifies that all system alerts correspond to actual fall events. Recall, on the other hand, measures the detection rate of all fall events; a score of 100.0\% in recall implies perfect detection. It is worth noting that in the machine learning community, ``recall" is more commonly used, while in medical testing and some other fields, ``Sensitivity" might be the preferred term to describe the same concept. For the sake of clarity and consistency, we will use the term ``recall" throughout the remainder of this paper. 

Specificity is another important measure and stands for the proportion of actual negatives that are correctly identified. It is defined as $Spec=\frac{TN}{TN+FP}$. While sensitivity/recall focuses on the correct identification of positive cases (falls in this context), Specificity ensures the correct identification of negative cases (non-fall activities). 

The F1 score, defined as $F1=2 \times \frac{Prec \times  Rec}{Prec+Rec}$, represents the harmonic mean of precision and recall. A perfect F1 score of 1.0 denotes perfect precision and recall. This score drops to 0 if either precision or recall is zero. 

\section{Radar Fall Detection Fundamentals}

\subsection{Radar Signal Processing}
In this section, we aim to provide a succinct overview of the fundamentals of radar signal processing. Our intention is to offer non-specialists in the radar domain a foundational understanding, even if it might be cursory in nature. For readers interested in a more comprehensive and in-depth exploration of radar signal processing, we recommend consulting the seminal book of Richards, Mark A.~\cite{richards2005fundamentals}. Radar, originally an acronym for ``Radio Detection and Ranging", has become a commonly used noun. Its applications can be broadly classified into detection, tracking, and imaging.

\subsubsection{Pulsed Radar}
In a pulsed radar system, a transmitter dispatches a pulse that is then reflected back to the receiver by an object. If the target is at distance $R$, the pulse traverses a total distance of $2R$. The required propagation time delay is denoted as $t_{0}$, and the distance to the target can thus be expressed as $R=\frac{ct_{0}}{2}$. Here, $c$ is the speed of electromagnetic wave propagation in free space. The maximum detectable distance by the radar, termed as the maximum unambiguous range $R_{max}$, corresponds to the furthest distance that a pulse can travel back and forth within the continuous transmission pulse interval $T$, also known as the pulse repetition interval (PRI). The pulse repetition frequency (PRF) is the reciprocal of $T$. Consequently, the maximum unambiguous range can be represented as $R_{max}=\frac{c\cdot PRI}{2}=\frac{c}{2\cdot PRF}$.

Pulsed radars offer several advantages, including long-range applications and the capacity to measure both range and velocity with ease. However, a significant downside is their requirement for high peak powers to ensure a satisfactory average power.

\subsubsection{Continuous Wave (CW) Radar}
\begin{figure}[htbp]
     \centering
     \begin{subfigure}[b]{0.2\textwidth}
         \centering
         \includegraphics[height=3cm]{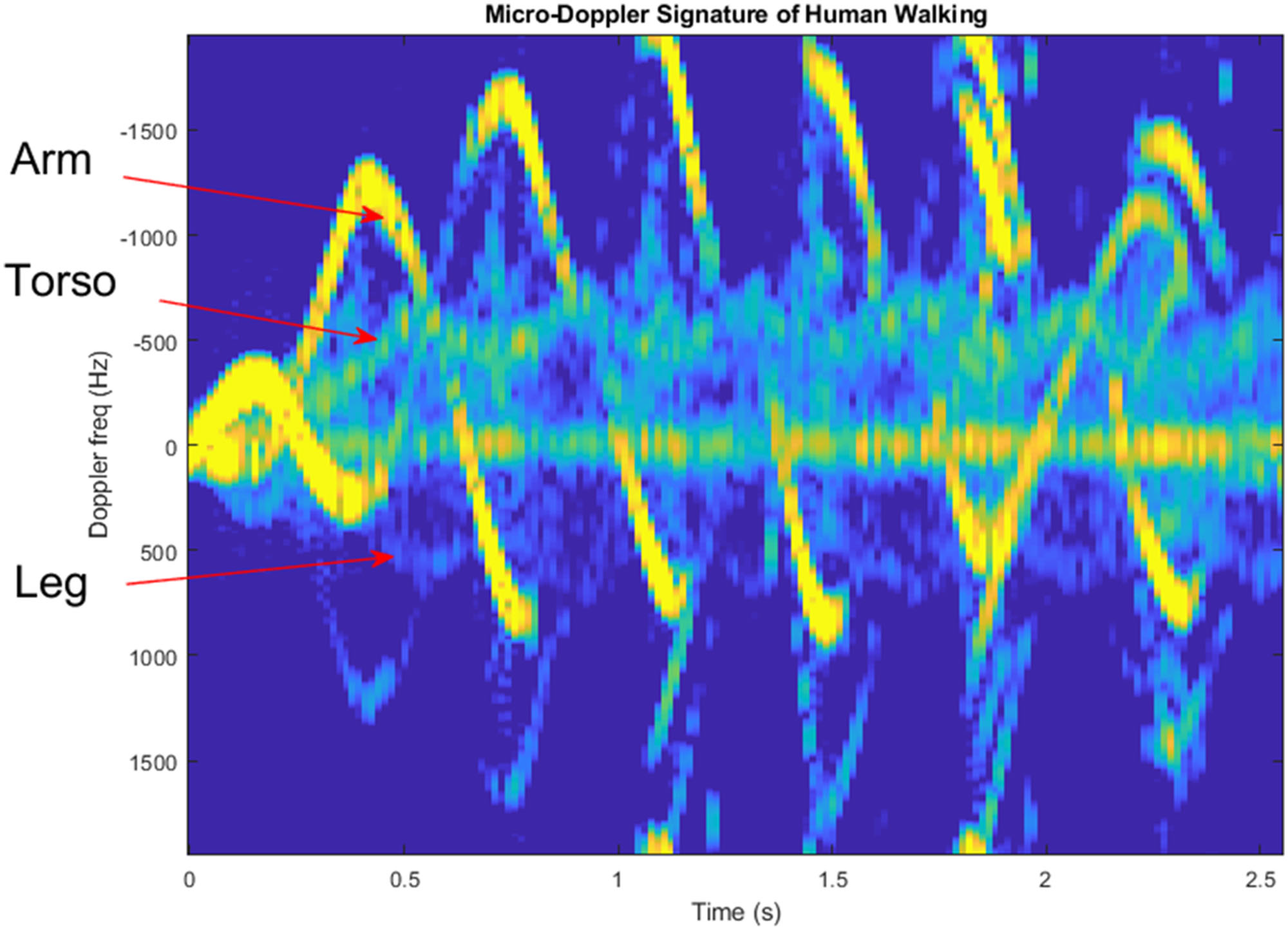}
         \caption{Micro-Doppler.~\cite{zhang2018real}}
         \label{fig:image1}
     \end{subfigure}
     \hfill
     \begin{subfigure}[b]{0.2\textwidth}
         \centering
         \includegraphics[height=3cm]{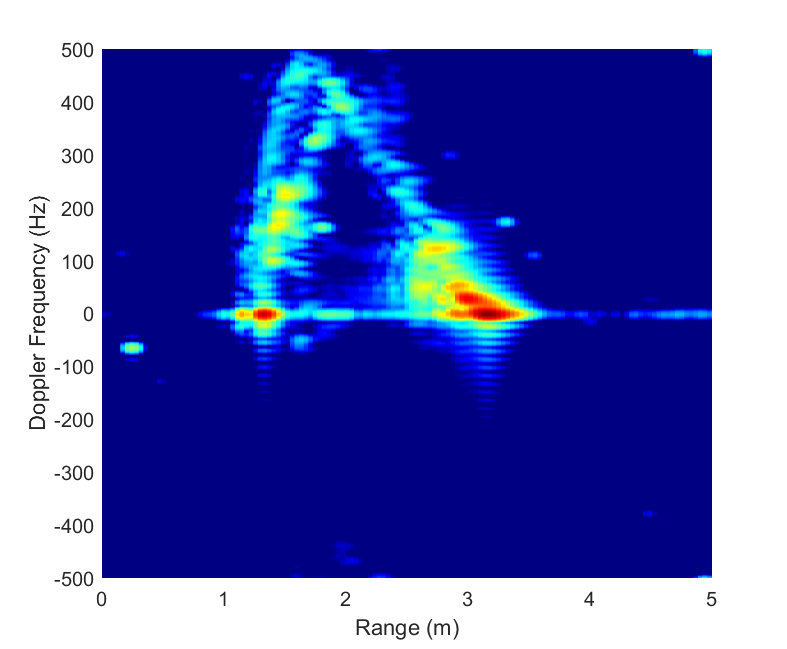}
         \caption{Range-Doppler.~\cite{erol2016wideband}}
         \label{fig:image2}
     \end{subfigure}
     \hfill
     \begin{subfigure}[b]{0.2\textwidth}
         \centering
         \includegraphics[height=3cm]{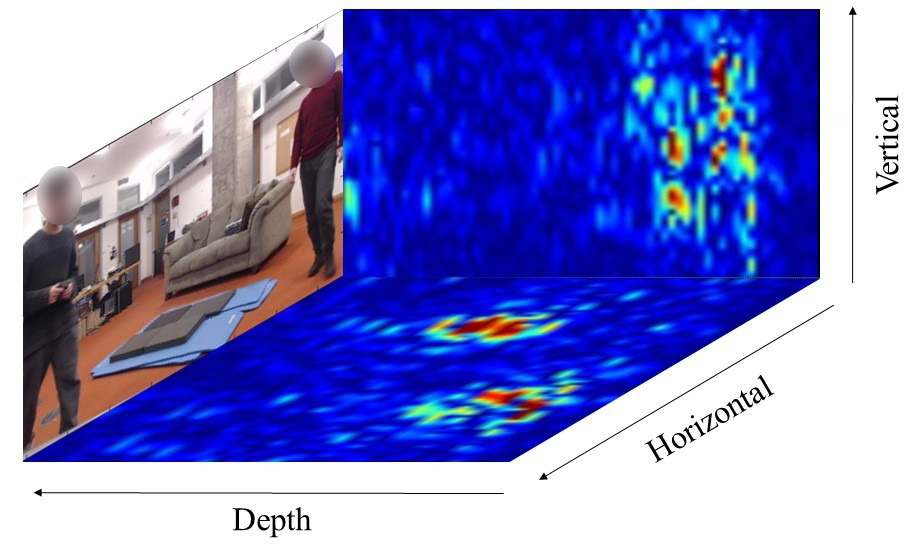}
         \caption{Range-Angles.~\cite{tian2018rf}}
         \label{fig:image3}
     \end{subfigure}
     \hfill
     \begin{subfigure}[b]{0.2\textwidth}
         \centering
         \includegraphics[height=3cm]{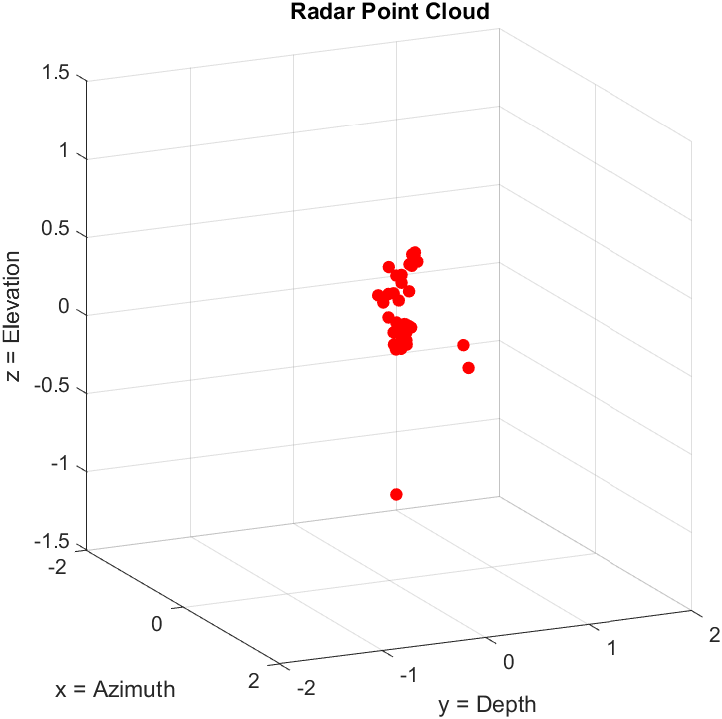}
         \caption{Point-Cloud.}
         \label{fig:image4}
     \end{subfigure}     
        \caption{Different radar data types of a human movements.}
        \label{fig:main}
\end{figure}

Unlike pulsed radar, CW radar transmits signals uninterruptedly, equating average power with peak power. CW radar listens for signal reflections from a target while transmitting, necessitating separate antennas for transmission and reception. A fixed frequency signal is transmitted, and the reflections from objects are received and mixed with the transmitted carrier. Like pulsed radar, CW radar detects the radial velocity of a moving object, which changes the frequency of the reflected signal. If the waveform is reflected from a target at distance $R_{0}$, moving at a constant velocity $v$ with a radial angle $\theta$, an arbitrarily time-varying range $R(t)$ can be defined as $R(t)=R_{0}+vcos(\theta)t$. The second component introduces a Doppler frequency shift in the radar return, which might consist of only a few Hz shifts on top of a multi-GHz carrier signal. In addition to overall movement, different parts of the target may also exhibit additional micro-scale movements, causing further Doppler shifts. These are known as micro-Doppler effects and can provide valuable information for identifying target characteristics. For instance, Figure~\ref{fig:image1}, the natural arm swing of a walking person generates a distinctive micro-Doppler effect. The stronger reflection from the human body, compared to the limbs, coupled with time-frequency representation of micro-Doppler effects, offers a wealth of information.

\subsubsection{Frequency Modulated Continuous Wave (FMCW) Radar}
\begin{figure}[htbp]
\centerline{\includegraphics[width=0.5\textwidth]{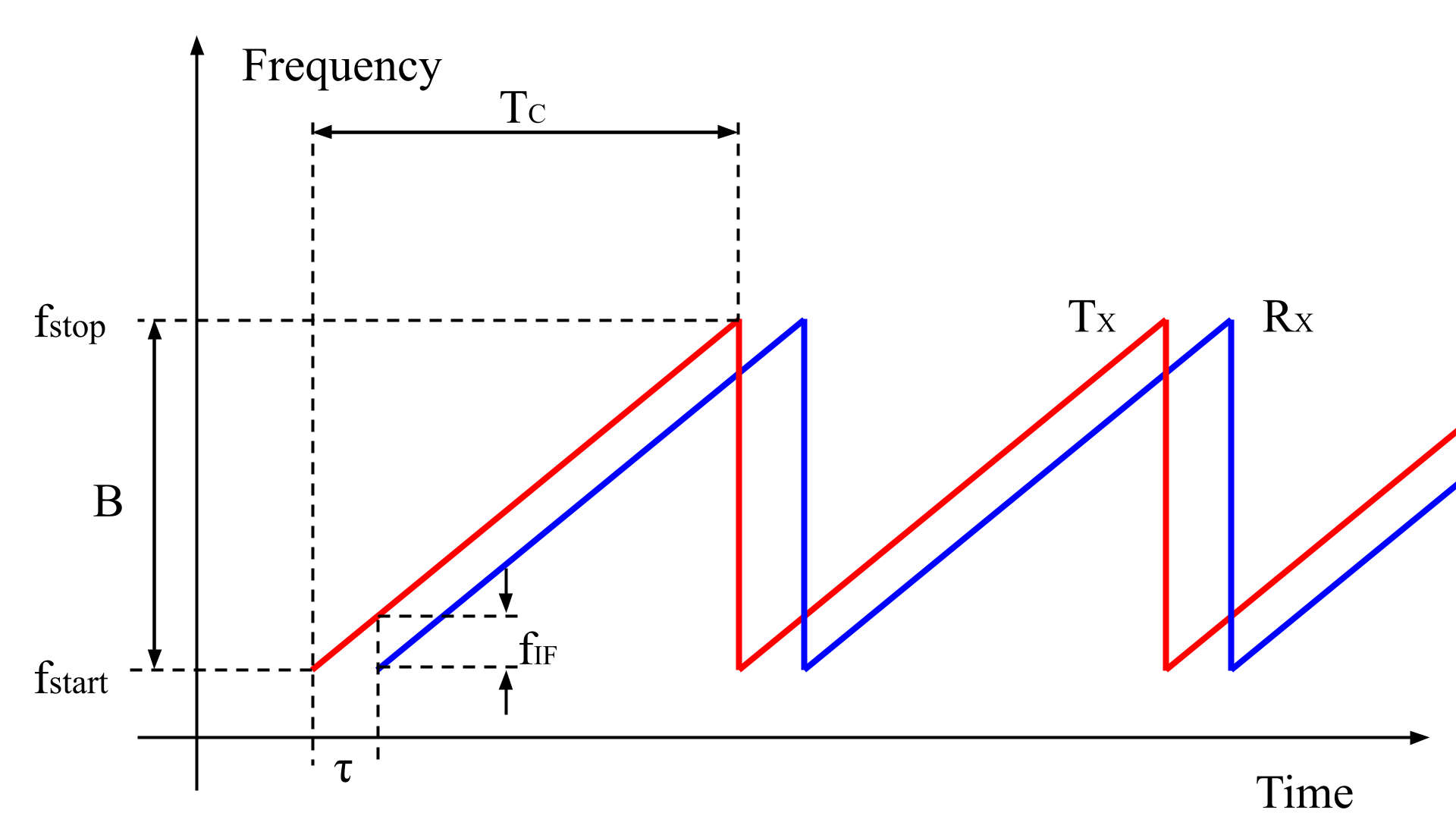}}
\caption{FMCW radar IF frequency signal.}
\label{fig5}
\end{figure}

While unmodulated CW radar can gauge velocity, it cannot measure target distance. To overcome this, FMCW radar transmits a frequency-modulated sinusoidal signal continuously to measure both range and velocity. In this case, the frequency increases linearly with time, generating a signal also known as a chirp. Figure~\ref{fig5} provides an example of a transmitted chirp signal and the reflection from a single detected object.

Similar to CW radar, FMCW radar requires a ``mixer" that combines the signals from the transmitter and receiver to produce an Intermediate Frequency (IF) signal. This frequency difference, $f_{IF}$, is known as the ``beat frequency". Its maximum value, the IF bandwidth, is dependent on the chip hardware performance. For example, the AWR2243 radar device from Texas Instruments (TI) offers a 20MHz bandwidth. Assuming the chirp's slope is $S=20MHz/us $, the ideal maximum range is: $R_{max}=\frac{c\cdot PRI}{2}=\frac{c}{2}\cdot \frac{f_{IF}}{S}=150m$. Given a target at distance $d$ from the radar, the IF signal will be a sinusoidal wave $x(t)=Asin(2\pi f_{0}t+\phi _{0})$.

The IF signal is valid from the moment the Receiver (RX) chirp arrives, so its initial phase $\phi _{0}$ is $\phi _{0}=2\pi f_{c}\tau $. Where $\tau$ is the time difference between when the Transmitter (TX) chirp is transmitted and RX chirp arrives, and $f_{0}=S\tau$. From $\tau =\frac{2d}{c}$ and $\lambda = c/f_{c}$, we can further deduce:

\begin{equation}
x(t)=Asin(2\pi \frac{S2d}{c} t+\frac{4\pi d}{\lambda })
\label{eq14}
\end{equation} 

Once the frequency $f_{0}=\frac{S2d}{c} $ of the sine wave at the moment of RX reception is known, we can calculate the distance $d$ to the target. This is applicable for a single target. For multiple targets, there will be multiple IF signals, which can be distinguished using the Fast Fourier Transform (FFT), referred to as range-FFT. However, FFT theory suggests that the smallest frequency component resolvable within an observation window $T_{c}$ is $\frac{1}{T_{c}}$. Given the bandwidth of FMCW radar is $B=f_{stop}-f_{start}=ST_{c}$, we can determine the minimum resolvable distance between two targets, or range resolution $R_{res}=\frac{c}{2B}$.

In Equation~\ref{eq14}, the second component $\phi _{0}=\frac{4\pi d}{\lambda }$ shows the change in phase. We can discover the velocity information of the target by sending two chirp signals with time interval $T_{c}$. The target moves a distance of $d=vT_{c}$ within $T_{c}$ time, and the phase change is $\Delta \phi =\frac{4\pi vT_{c}}{\lambda }$. Once we know the phase information, we can calculate the velocity of the target $v=\frac{\lambda \Delta \phi }{4\pi T_{c}}$. But since the measurement of velocity is based on the phase difference, the maximum unambiguous phase difference is $\pi$, so our maximum unambiguous velocity $v_{max}=\frac{\lambda }{4T_{c}}$.

To differentiate multiple targets moving at the same distance simultaneously, we need to execute another FFT across multiple consecutive chirps, known as velocity-FFT. We define the spatial frequency $\omega = \Delta \phi = \frac{4\pi vT_{c}}{\lambda }$. The term spatial frequency refers to the phase-shift across consecutive chirps. Suppose we have two targets and the velocity difference is $\Delta v $. The difference in the spatial frequency corresponding to these two objects is $\Delta \omega = \frac{4\pi T_{c}}{\lambda }\Delta v $. The theory of discrete Fourier transforms tells us that within N consecutive chirps cycles, the smallest spatial frequency component that can be resolved is $\frac{2\pi }{N}$. We can get the minimum velocity resolution $v_{res}=\frac{\lambda }{2NT_{c}}$. N consecutive chirps is called a frame, we define frame time $T_{f}=NT_{c}$, then $v_{res}=\frac{\lambda }{2T_{f}}$. 

In order to get the angle of the target relative to the radar, which is also called the angle of arrival (AoA), we need at least two RX antennas. We define the distance between two adjacent RX antennas as $d$, and the incident signal reflected from the target reaches each antenna with a distance difference of $dsin\theta $. We can do an angle-FFT along the antenna dimension $\{1, 2, ..., N\}$, and the differential distance result in a phase change in the FFT peak. The phase change from the target to the RX antenna can be expressed as $\phi =2\pi f_{c}t=2\pi \frac{c}{\lambda }\frac{d}{c}=\frac{2\pi d}{\lambda }$. The phase difference due to $\Delta d=dsin\theta $ is $\Delta \phi =\frac{2\pi dsin\theta}{\lambda }$. We can get $\theta =sin^{-1}(\frac{\lambda \Delta \phi }{2\pi d})$. Similarly, the maximum unambiguous phase difference is $\pi $, so our maximum unambiguous angle $\theta _{max}=sin^{-1}(\frac{\lambda }{2d})$. We usually choose $d=\lambda /2$ when designing antennas, because then we can get the largest field of view (FoV) $\theta _{max}\in (-90^{\circ }, 90^{\circ })$. 

The angle resolution $\theta _{res}$ depends on the number of receiver antennas available. The larger the number of antennas, the better the resolution. Again, we let the spatial frequency $\omega = \Delta \phi=\frac{2\pi dsin\theta}{\lambda }$. The term spatial frequency here refers to the phase shift across consecutive antennas in the RX array. Suppose we have two targets and the AoA difference is $\Delta \theta$. The difference in the spatial frequency corresponding to these two objects $\Delta \omega  =\frac{2\pi d}{\lambda }(sin(\theta + \Delta \theta)-sin\theta)\approx \frac{2\pi d}{\lambda}(cos\theta \Delta \theta)$. Similarly, we need $\Delta \omega \geq  \frac{2\pi }{N}$ and can get the the minimum angle resolution $\theta _{res}=\frac{\lambda }{dNcos\theta }=\frac{2}{N},\ when\ d=\lambda /2$. 

\subsection{Machine Learning for Fall Detection}
In recent years, machine learning (ML) has gained prominence as an instrumental technique in healthcare, especially in fall detection and prediction among the elderly population~\cite{singh2022systematic}. While traditional fall detection mechanisms predominantly utilize threshold-based strategies, they exhibit constraints concerning adaptability and precision~\cite{bourke2008threshold}. In contrast, the data-driven nature of ML models makes them adapt better to various scenarios and individual fall patterns. Deep Learning (DL), outside of ML, is another subset of Artificial Intelligence (AI) and forms the computational foundation for fall detection. ML empowers computers to acquire knowledge and execute tasks without being explicitly programmed~\cite{zhou2021machine}~\cite{jordan2015machine}. DL, a more specialized category within ML, builds on this by leveraging layered architectures that emulate neural structures in the human brain, enabling non-linear data transformations~\cite{lecun2015deep}~\cite{goodfellow2016deep}.

\subsubsection{Machine Learning for Fall Detection}
Traditional ML techniques, such as Naive Bayes (NB) Classifier~\cite{xu2018bayesian}, Support Vector Machine (SVM)~\cite{pisner2020support}, Decision Trees (DT)\cite{izza2020explaining}, K-means \& k-Nearest Neighbors (kNN)\cite{mittal2019performance}, and Principal Component Analysis (PCA)~\cite{xie2019principal}, have served as foundational methodologies in fall detection research. These algorithms, widely recognized for their interpretability and efficiency, have provided reliable results when applied to structured datasets. Additionally, they often require manual feature extraction, where domain knowledge plays a crucial role in achieving optimal performance. The choice of a specific ML technique often stems from the nature of the data at hand, the desired feature representation, and the underlying problem's complexity. Furthermore, the ease of implementation and interpretability of traditional ML techniques have made them popular in scenarios where understanding the decision-making process is pivotal.

\subsubsection{Deep Learning for Fall Detection}
\begin{figure}[htbp]
\centerline{\includegraphics[width=0.5\textwidth]{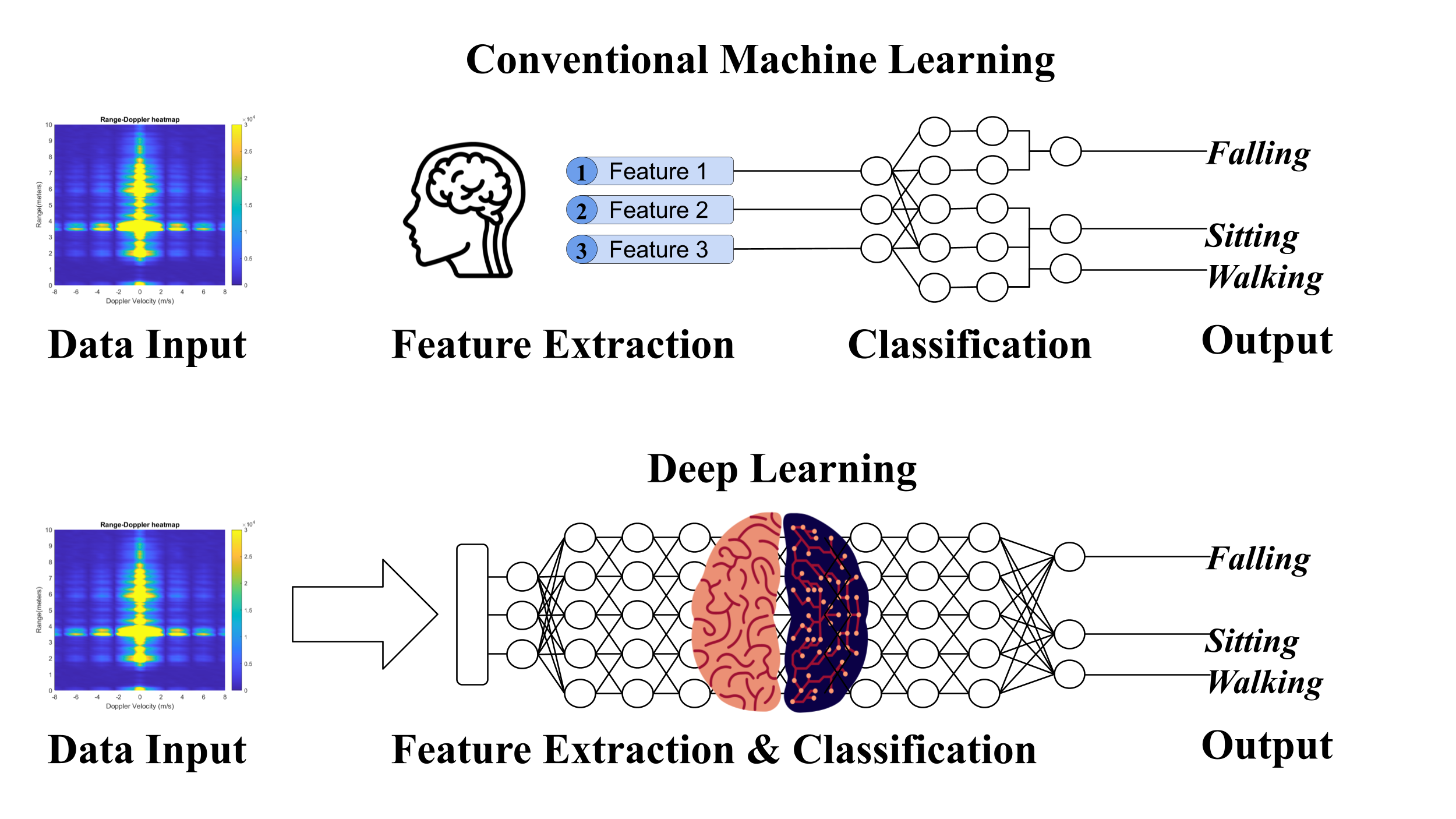}}
\caption{Distinguishing Machine Learning and Deep Learning.}
\label{fig8}
\end{figure}

Emerging as the vanguard of modern computational approaches, deep learning architectures, such as Convolutional Neural Networks (CNNs)~\cite{krizhevsky2012imagenet}, Recurrent Neural Networks (RNNs)~\cite{yu2019review}, Variational Autoencoders (VAEs)\cite{doersch2016tutorial}, Generative Adversarial Networks (GANs)~\cite{aggarwal2021generative}, and Adversarial Autoencoders (AAEs)~\cite{makhzani2015adversarial}, have marked a transformative shift in fall detection. Their unique ability to automatically learn hierarchical features from raw data, coupled with their proficiency in handling large volumes of unstructured data, make them particularly suited for complex tasks like analyzing radar signals. These deep architectures have been pivotal in bypassing manual feature engineering, often a cumbersome step in traditional machine learning. As such, they present a more holistic approach, learning both low-level and high-level features that are essential for tasks. The choice of a specific DL architecture is primarily influenced by the dataset's complexity, the nature of the task, and the available computational power. Their adaptability, scalability, and proficiency in capturing intricate patterns are instrumental in pushing the boundaries of what's achievable in fall detection. Figure~\ref{fig8} shows us the difference between Machine Learning and Deep Learning.

CNNs exploit local data dependencies to minimize learnable weights, facilitating the construction of deeper networks that recognize intricate features. 1D-CNNs are suitable for time-series data, with kernels sliding along one dimension, treating different time-series as distinct filter channels. 2D-CNNs, on the other hand, are optimal for data with multiple features like image, making them apt for processing radar-derived data such as Micro-Doppler and Range-Doppler maps. Lastly, 3D-CNNs are best suited for multidimensional data like radar point clouds, where features such as Doppler and/or Signal-to-noise ratio (SNR) serve as unique filter channels.

Human activities are performed continuously over time, making the temporal dimension crucial for understanding behavior. While RNNs excel at processing sequential data, they suffer from short-term memory limitations. Advanced variants like Long Short-Term Memory (LSTM) and Gated Recurrent Units (GRU) address this by learning long-term dependencies through additional gates. While RNNs handle hierarchical tree structures, the sequence-to-sequence (Seq2Seq) model, using encoders and decoders, struggles with longer sequences. The attention mechanism addresses this by weighting input significance, and transformers eschew RNNs altogether, focusing on multi-head attention. Moreover, combining RNNs with CNNs facilitates spatio-temporal processing of data. 

Autoencoders, including VAEs, compress data into a latent space to uncover underlying features, and can reconstruct or even generate new data samples. VAEs particularly use Bayesian statistics to learn data distributions. On the other hand, GANs employ a generator and discriminator to produce data samples without assuming a specific distribution, improving through iterative training until the generated samples are indistinguishable from real ones. AAEs merge VAE and GAN concepts, leveraging adversarial loss for flexible distribution choices in the latent space, thus comprising an encoder-generator, decoder, and discriminator setup. For fall detection applications, especially when labeling data is time-consuming, one can consider employing unsupervised/semi-supervised methods using VAEs, GANs, or AAEs. 

\section{Review of Radar-Based Fall Detection} \label{Review}

This section presents a comprehensive review of 74 studies focused on radar-based fall detection, which utilizes different radar data formats such as Micro-Doppler, Range-Doppler, and Range-Doppler-Angles, exemplified in Figure~\ref{fig:main}. The reviewed researches are systematically classified into three distinct groups based on their unique characteristics. Table~\ref{tab4} offers a clear categorization of these studies, organizing them according to the specific classifiers employed within each category. It is important to note that for studies that utilized multiple classifiers or algorithms, only the one that reported the best performance is showcased, unless the authors explicitly indicate a combination of methods, e.g., ~\cite{amin2016radar}~\cite{erol2017range}~\cite{erol2019radar}.

\begin{table*}[!htbp]
\renewcommand{\arraystretch}{1.3}
\caption{Categorization of reviewed publications based on radar data formats and the classifiers}
\label{tab4}
\centering
\scriptsize
\begin{threeparttable}
\begin{tabularx}{1.0\textwidth}{|X|l|l|l|l|l|l|l|l|l|l|l|l|l|}
\hline
Category &
  Threshold &
  SVM &
  Bayes\tnote{1} &
  kNN\tnote{2} &
  LDA\tnote{3} &
  SVDD\tnote{4} &
  PCA &
  BPNN\tnote{5} &
  CNN\tnote{6} &
  RNN\tnote{7} &
  AE\tnote{8} &
  GAN &
  Other\tnote{9} \\ \hline
Micro-Doppler &
  ~\cite{amin2015personalized} &
  \begin{tabular}[c]{@{}l@{}}~\cite{liu2012doppler}\\ ~\cite{karsmakers2012automatic}\\ ~\cite{hong2013cooperative}\\ ~\cite{wu2013fall} \\ ~\cite{garripoli2014embedded}\\ ~\cite{jokanovic2015multi}\\ ~\cite{amin2016radar} \\ ~\cite{liu2016automatic}\\ ~\cite{shrestha2017feature}\\ ~\cite{li2017multisensor} \\ ~\cite{li2019activities} \\ ~\cite{alnaeb2019detection} \\ ~\cite{hanifi2021elderly} \\ ~\cite{liu2014automatic}\end{tabular} &
  ~\cite{wu2015radar} &
  \begin{tabular}[c]{@{}l@{}}~\cite{liu2011automatic}\\ ~\cite{liu2012fall}\\ ~\cite{su2014doppler}\\ ~\cite{amin2016radar} \\ ~\cite{su2018radar}\\ ~\cite{erol2018realization} \\ ~\cite{sadreazami2020compressed}\end{tabular} &
   &
  ~\cite{chen2022three} &
  ~\cite{jokanovic2016radarPCA} &
   &
  \begin{tabular}[c]{@{}l@{}}~\cite{zhou2018fall}\\ ~\cite{sadreazami2019fall}\\ ~\cite{sadreazami2019tl} \\ ~\cite{sadreazami2019residual} \\ ~\cite{sadreazami2019capsfall} \\ ~\cite{yoshino2019fall} \\ ~\cite{anishchenko2019fall}\\ ~\cite{li2020distributed}\\ ~\cite{chuma2020internet} \\ ~\cite{maitre2020fall} \\ ~\cite{sadreazami2021contactless} \\ ~\cite{wang2022convolution}\\ ~\cite{wang2022millimeter}\\ ~\cite{saho2022machine}\end{tabular} &
  \begin{tabular}[c]{@{}l@{}}~\cite{sadreazami2018use}\\ ~\cite{li2019bi}\\ ~\cite{imamura2022automatic}\\ ~\cite{maitre2020fall}\end{tabular} &
  \begin{tabular}[c]{@{}l@{}}~\cite{jokanovic2016radarDL} \\ ~\cite{seyfiouglu2018deep} \\ ~\cite{shah2022data}\end{tabular} &
  \begin{tabular}[c]{@{}l@{}}~\cite{lu2021radar}\end{tabular} &
  \begin{tabular}[c]{@{}l@{}}~\cite{gadde2014fall}\\ ~\cite{rivera2014radar}\\ ~\cite{chen2022three}\end{tabular} \\ \hline
Range-Doppler &
   &
  \begin{tabular}[c]{@{}l@{}}~\cite{erol2016wideband} \\ ~\cite{erol2016fall}\\ ~\cite{erol2016range} \\ ~\cite{erol2017range}\\ ~\cite{ding2021sparsity}\end{tabular} &
   &
  \begin{tabular}[c]{@{}l@{}}~\cite{erol2017range}\\ ~\cite{erol2018radar} \\ ~\cite{ding2019fall} \\ ~\cite{ding2019continuous}\end{tabular} &
  ~\cite{erol2019radar} &
   &
  ~\cite{erol2019radar} &
  \begin{tabular}[c]{@{}l@{}}~\cite{he2019human} \\ ~\cite{erol2019radar}\end{tabular} &
  \begin{tabular}[c]{@{}l@{}}~\cite{shankar2019radar}\\ ~\cite{wang2020millimetre}\\ ~\cite{ma2020room}\\ ~\cite{bhattacharya2020deep}\end{tabular} &
  \begin{tabular}[c]{@{}l@{}}~\cite{liang2021posture} \\ ~\cite{ma2020room}\end{tabular} &
  ~\cite{jokanovic2017fall} &
   &
   \\ \hline
Range-Doppler-Angles &
   &
   &
   &
  ~\cite{ding2021fall} &
   &
   &
   &
  ~\cite{liang2022fall} &
  \begin{tabular}[c]{@{}l@{}}~\cite{tian2018rf} \\ ~\cite{yao2022fall}\end{tabular} &
  ~\cite{sun2019privacy} &
  ~\cite{jin2020mmfall} &
   &
   \\ \hline
\end{tabularx}
\begin{tablenotes}
\item[1] Naive Bayes, or Sparse Bayesian classifier.
\item[2] K-means, or k-Nearest Neighbors.
\item[3] Linear discriminant analysis (LDA).
\item[4] Support Vector Data Description (SVDD).
\item[5] Back Propagation Neural Network (BPNN), Shallow Neural Networks.
\item[6] Deep CNNs, and variants including ResNet, AlexNet, VGG, Inception, DenseNet, etc.
\item[7] Deep RNNs, and variants including LSTM, GRU, Seq2Seq, Attention, Transformer, etc.
\item[8] Including AE, VAE, AAE.
\item[9] Quadratic Discriminant Analysis (QDA), Adaptive Boosting (AdaBoost), Decision Tree, Logistic Regression (LR), Mahalanobis Distance classifier.
\end{tablenotes}
\end{threeparttable}
\end{table*}

\begin{figure}[htbp]
\centerline{\includegraphics[width=0.45\textwidth]{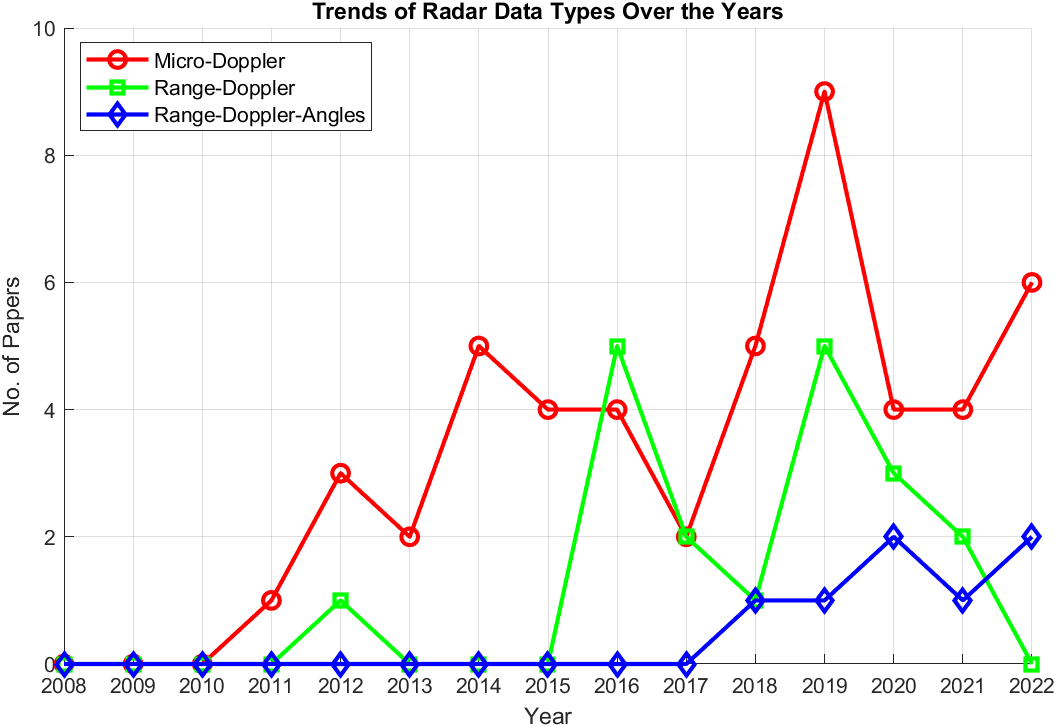}}
\caption{Temporal Evolution of Radar Data Types Utilized in Fall Detection Research.}
\label{fig4}
\end{figure}

Early studies in fall detection primarily leveraged Micro-Doppler data. However, with the introduction of FMCW radar technology, research incorporating range information began emerging around 2016. While Micro-Doppler remains a leading method in fall detection, advancements in Printed Circuit Board (PCB) onboard antennas and the evolution of Multiple-Input Multiple-Output (MIMO) technology have steered a rising number of studies towards using Range-Doppler-Angles data since 2018. The progression of radar data types employed in fall detection over the years can be visualized in Figure~\ref{fig4}.

While there is extensive research on radar-based human-related applications, such as human activity classification~\cite{kim2009human}~\cite{zhang2018real}~\cite{jin2019multiple}, and pose estimation~\cite{sengupta2020mm}~\cite{sengupta2020nlp}, this review does not encompass those studies. Our focus is specifically on research that addresses fall detection as the primary problem, hence, the reviewed papers must either contain the keyword ``fall" or primarily tackle the issue of fall detection.
    
\subsection{Fall Detection with Micro-Doppler}

Micro-Doppler-based fall detection has seen significant advancements over the years, all centered around the correlation between Doppler frequency and motion velocity. 

Liu, Liang, et al. laid the groundwork, using Mel-Frequency Cepstral Coefficients (MFCC) to represent Doppler signatures of human activities, employing SVM and kNN classifiers for fall detection~\cite{liu2011automatic}~\cite{liu2012fall}~\cite{liu2012doppler}. Karsmakers, Peter, et al.~\cite{karsmakers2012automatic} improved accuracy with a CW Doppler radar and the Global Alignment (GA) kernel. Hong, Jihoon, et al.~\cite{hong2013cooperative} addressed non-line-of-sight (NLOS) effects, while Liu, Liang, et al.~\cite{liu2014automatic} developed practical applications for senior apartments.

Later studies focused on feature extraction techniques and classifiers. Gadde, Ajay, et al.~\cite{gadde2014fall} used Time-Scale-based signal analysis and the Mahalanobis Distance (MD) classifier, while Su, Bo Yu, et al.~\cite{su2014doppler}~\cite{su2018radar} combined Wavelet Transform (WT) and a nearest neighbor classifier for improved results.

Deep learning entered the scene with Jokanovic, Branka, et al.~\cite{jokanovic2016radarPCA}~\cite{jokanovic2016radarDL} employing PCA and stacked Autoencoders. Amin, Moeness G. et al.~\cite{amin2016radar}~\cite{amin2017radar} advocated for elderly-specific radar monitoring algorithms and a larger fall data repository. The use of CNNs gained traction with Zhou et al.~\cite{zhou2018fall} and Yoshino et al.~\cite{yoshino2019fall} achieving high accuracies. Seyfio{\u{g}}lu, Mehmet Sayg{\i}n, et al.~\cite{seyfiouglu2018deep} implemented a deep Convolutional Autoencoder (CAE) for classifying micro-Doppler signatures. Sadreazami, Hamidreza, et al.~\cite{sadreazami2018use}~\cite{sadreazami2019fall}~\cite{sadreazami2019tl}~\cite{sadreazami2019residual}~\cite{sadreazami2019capsfall}~\cite{sadreazami2020compressed}~\cite{sadreazami2021contactless} experimented with various network architectures, while Anishchenko et al.~\cite{anishchenko2019fall} improved reliability with two radars at a 90° angle.

Recent work by Lu et al.~\cite{lu2021radar} and Chen et al.~\cite{chen2022three} explored three-stage fall detection approaches, reducing power consumption and improving detection. Studies by Wang et al.~\cite{wang2022convolution}~\cite{wang2022millimeter} focused on signal processing and soft-fall detection. Shah et al.~\cite{shah2022data} achieved an accuracy of 88.0\% using an autoencoder. The latest research by Saho et al.~\cite{saho2022machine} used two Doppler radars installed in a restroom, highlighting the continuous evolution of Micro-Doppler-based fall detection, moving from foundational exploration towards practical, real-world applications with improved accuracy.
    
Micro-Doppler-based fall detection has progressed from using basic classifiers to advanced deep learning models, enhancing accuracy and practicality. The work began with SVM and kNN classifiers, and has evolved to include the use of CNNs and GANs.
Sensor configurations and fusion techniques have improved system performance, with multiple radars and sensor types reducing false alarms. Emphasizing real-world applications, like systems for senior apartments, demonstrates the field's maturity and readiness for deployment.
Despite challenges, like NLOS effects and data distortion, the advancements in this field are promising for creating robust, efficient fall detection systems. Future research should focus on these issues while exploring new deep learning techniques and sensor configurations.
The summary table for Fall Detection with Micro-Doppler is Table~\ref{tab5}. 

Please note that, in our discussion in Section~\ref{Quality Metrics}, we emphasize the importance of precision, recall, and the F1 score as performance metrics for fall detection systems. However, the metrics reported in the reviewed papers varied. To make the performance metrics consistent (i) For papers with a provided confusion matrix, we computed the precision, recall, and F1 score directly; (ii) In cases where sensitivity was mentioned, we treated it synonymously with recall while opting to overlook the specificity; (iii) When precision and false alarm rate were present, we deduced the recall and subsequently calculated the F1 score; (iv) For instances where only accuracy (most frequently in classification tasks where a fall is one among multiple classes) or the Area Under the Receiver Operating Characteristics Curve (AUC) was reported, we maintained the results as presented in the original studies. Even though we make these efforts, inconsistencies in Table~\ref{tab5}~\ref{tab6}~\ref{tab7} are still inevitable. We believe that with the broader adoption of standardized metrics in the field, such inconsistencies will be reduced in future research. 
    
\begin{table*}[htbp]
\renewcommand{\arraystretch}{1.3}
\caption{Fall Detection with Micro-Doppler}
\label{tab5}
\centering
\scriptsize
\definecolor{headercolor}{rgb}{0.88,0.88,0.88} 
\begin{tabularx}{1.0\textwidth}{|p{3cm}|p{1.5cm}|p{1.5cm}|p{1cm}|p{4cm}|X|}
\hline
\rowcolor{headercolor} \textbf{Paper} & \textbf{Radar} & \textbf{Classifier} & \textbf{Subjects} & \textbf{Motions} & \textbf{Performance}\\
\hline
Liu, Liang, et al.~\cite{liu2011automatic} & Doppler & kNN & 3 & 109 falls, 341 non-falls & AUC=0.974\\
\hline
Liu, Liang, et al.~\cite{liu2012fall} & Doppler & kNN & 3 & 109 falls, 341 non-falls & AUC=0.979\\
\hline
Liu, Liang, et al. ~\cite{liu2012doppler} & Doppler & SVM & 2 & 216 falls, 1158 non-falls & AUC=0.996\\
\hline
Karsmakers, Peter, et al.~\cite{karsmakers2012automatic} & Doppler & SVM & 2 & 4 activities, 60 examples & Acc=95.0\%\\
\hline
Hong, Jihoon, et al.~\cite{hong2013cooperative} & Doppler & SVM & - &  4 activities, 80 examples & AUC=0.964\\
\hline
Wu, Meng, et al.~\cite{wu2013fall}  & Doppler & SVM & - & 3 activities, 30 examples & Acc=100.0\%\\
\hline
Liu, Liang, et al.~\cite{liu2014automatic} & Doppler & SVM & 1 & 72 falls, 98 non-falls & AUC=0.980\\
\hline
Gadde, Ajay, et al.~\cite{gadde2014fall} & Doppler & MD & 2 & 10 falls, 10 non-falls & Prec=100.0\%, Rec=100.0\%, F1=100.0\%\\
\hline
Rivera, Luis Ramirez, et al.~\cite{rivera2014radar} & Doppler & MD & 2 & 8 activities, 80 examples & Prec=100.0\%, Rec=97.5\%, F1=98.7\%\\
\hline
Garripoli, Carmine, et al.~\cite{garripoli2014embedded} & Doppler & SVM & 16 & 4 activities, 65 falls & Prec=100.0\%, Rec=100.0\%, F1=100.0\%\\
\hline
Su, Bo Yu, et al.~\cite{su2014doppler} & Doppler & kNN & 2 & 105 falls, 990 non-falls & AUC=0.960, Rec=97.1\% \\
\hline
Wu, Qisong, et al.~\cite{wu2015radar} & Doppler & Bayes & 2 & 8 activities, 80 examples & -\\
\hline
Jokanovic, Branka, et al.~\cite{jokanovic2015multi} & Doppler & SVM & 2 & 5 activities, 32 falls and 32 non-falls & AUC=0.938\\
\hline
Amin, Moeness, et al.~\cite{amin2015high} & Doppler  & - & - & 20,000 samples & -\\
\hline
Amin, Moeness, et al. ~\cite{amin2015personalized}  & Doppler & Threshold & 5 & 20,000 samples & Rec=87.5\%\\
\hline
Jokanovic, Branka, et al.~\cite{jokanovic2016radarPCA}  & Doppler & PCA & - & 4 activities, 60 examples & Acc=90.0\%\\
\hline
Jokanovic, Branka, et al.~\cite{jokanovic2016radarDL}  & Doppler & AE & - & 4 activities, 120 examples & Acc=87.0\%\\
\hline
Amin, Moeness, et al.~\cite{amin2016radar}  & Doppler & SVM, kNN & 1 & 19 types of falls, 14 types of non-falls & Prec=100.0\%, Rec=82.0\%\\
\hline
Liu, Liang, et al.~\cite{liu2016automatic} & Doppler & SVM & 6 & real-life senior resident activities & 2.0/week false alarms, Rec=100.0\%\\
\hline
Shrestha, A., et al.~\cite{shrestha2017feature} & FMCW & SVM & 6 & 7 activities & Acc=94.0\%\\
\hline
Li, Haobo, et al.~\cite{li2017multisensor}  & FMCW & SVM & 16 & 10 activities & Acc=91.3\%\\
\hline
Su, Bo Yu, et al.~\cite{su2018radar} & Doppler & kNN & 1  & real-life senior resident activities & 8.6/day false alarms, Rec=100.0\%\\
\hline
Seyfio{\u{g}}lu, Mehmet Sayg{\i}n, et al. ~\cite{seyfiouglu2018deep}  & Doppler & CAE & 11 & 12 activities & Acc=94.2\%\\
\hline
Sadreazami, Hamidreza, et al.~\cite{sadreazami2018use} & UWB & LSTM & 5 & 121 falls, 85 non-falls & Prec=95.0\%, Rec=88.5\%, F1=91.6\%\\
\hline
Erol, Baris, et al.~\cite{erol2018realization}  & Doppler & PCA  & 14 & 6 activities & Rec=97.0\%\\
\hline
Zhou, Xu, et al.~\cite{zhou2018fall} & Doppler & CNN & 3 & 4 activities & Acc=99.85\%\\
\hline
Sadreazami, Hamidreza, et al.~\cite{sadreazami2019fall} & UWB & CNN & 5 & 121 falls, 85 non-falls & Prec=94.2\%, Rec=93.4\%, F1=93.8\%\\
\hline
Sadreazami, Hamidreza, et al.~\cite{sadreazami2019tl}  & UWB & CNN & 5 & 121 falls, 85 non-falls & Prec=96.1\%, Rec=96.7\%, F1=96.4\%\\
\hline
Sadreazami, Hamidreza, et al.~\cite{sadreazami2019residual} & UWB & CNN & 10 & 187 falls, 149 non-falls & Prec=96.2\%, Rec=90.9\%, F1=93.5\%\\
\hline
Sadreazami, Hamidreza, et al.~\cite{sadreazami2019capsfall}  & UWB & CNN & 10 & 1870 falls, 1490 non-falls & Prec=93.6\%, Rec=90.4\%, F1=92.0\%\\
\hline
Yoshino, Haruka, et al. ~\cite{yoshino2019fall} & Doppler & CNN & 10 & 2 types of falls, 2 types of non-falls & Prec=95.0\%\\
\hline
Li, Haobo, et al.~\cite{li2019activities}  & FMCW & SVM & 16 & 6 activities & Acc=87.3\%\\
\hline
Anishchenko, Lesya, et al.~\cite{anishchenko2019fall} & Doppler & CNN & 5 & 175 falls, 175 non-falls & Prec=100.0\%, Rec=98.6\%, F1=99.3\%\\
\hline
Alnaeb, Ali, et al.~\cite{alnaeb2019detection}  & Doppler & SVM & 5 & 50 falls, 50 non-falls & Acc=100.0\%\\
\hline
Li, Haobo, et al.~\cite{li2019bi} & FMCW & LSTM & 16 & 6 activities & Acc=96.0\%\\
\hline
Li, Haobo, et al.~\cite{li2020distributed} & FMCW & CNN & 14 & 12 activities & Acc=84.0\%\\
\hline
Sadreazami, Hamidreza, et al.~\cite{sadreazami2020compressed}  & UWB & kNN & 5 & 6 activities & Prec=94.3\%, Rec=100.0\%, F1=97.1\%\\
\hline
Chuma, Euclides Lourenço, et al.~\cite{chuma2020internet}  & Doppler & CNN & 6 & 5 activities & Acc=99.9\%\\
\hline
Maitre, Julien, et al.~\cite{maitre2020fall}  & UWB & CNN-LSTM & 10 & 60 falls, 600 non-falls & Acc=90.0\%\\
\hline
Lu, Jincheng, et al.~\cite{lu2021radar} & Doppler & GAN & 3 & 14 activities & Prec=91.8\%, Rec=93.3\%, F1=92.6\%\\
\hline
Sadreazami, Hamidreza, et al.~\cite{sadreazami2021contactless}  & UWB  & CNN & 10 & 187 falls, 149 non-falls & Prec=98.4\%, Rec=94.4\%, F1=96.28\%\\
\hline
Hanifi, Khadija, et al.~\cite{hanifi2021elderly} & Doppler & SVM & 10 & 543 falls and 675 non-falls & Prec=88.0\%, Rec=87.0\%, F1=88.0\%\\
\hline
Chen, Mengxia, et al.~\cite{chen2022three} & UWB & SVDD, MD & 11 & 9 activities & Prec=99.6\%, Rec=97.6\%, F1=98.6\%\\
\hline
Wang, Ping, et al.~\cite{wang2022convolution} & UWB & CNN & 9 & 3 activities & Prec=91.7\%, Rec=95.7\%, F1=91.9\%\\
\hline
Shah, Syed Aziz, et al.~\cite{shah2022data}  & FMCW & AE & 99 & 6 activities & Acc=88.0\%\\
\hline
Wang, Bo, et al.~\cite{wang2022millimeter} & FMCW & CNN & 12 & 29 activities & Rec=95.8\%\\
\hline
Imamura, Takayuki, et al.~\cite{imamura2022automatic}  & Doppler & LSTM & 2 & 6 activities & Prec=95.2\%, Rec=96.6\%, F1=95.9\%\\
\hline
Saho, Kenshi, et al.~\cite{saho2022machine} & Doppler & CNN & 21 & 8 activities & Acc=95.6\%\\
\hline
\end{tabularx}

\begin{flushleft}
\textbf{Note:}
\begin{itemize}
\item \textbf{Acc (Accuracy)}: Represents the proportion of samples that are correctly predicted out of all the samples.
\item \textbf{Prec (Precision or Positive Predictive Value)}: Out of the samples predicted as positive, the proportion that are actually positive. 
\item \textbf{Rec (Recall or Sensitivity or True Positive Rate)}: The proportion of actual positive samples that are correctly predicted as positive by the model. 
\item \textbf{AUC}: Represents the Area Under the receiver operating characteristics (ROC) Curve. 
\end{itemize}
\end{flushleft}

\end{table*}

\subsection{Fall Detection with Range-Doppler}

Radar signals in the Time-Frequency (TF) domain reveal velocities, accelerations, and Doppler terms of human body parts in motion~\cite{mercuri2012sfcw}. These signals, combined with range information, allow for accurate positioning and movement tracking. 

Stepped-Frequency Continuous Wave (SFCW) radar, fixed both on the wall and on the ceiling, measures position and speed and has been deemed suitable for fall detection and vital signs detection~\cite{mercuri2012sfcw}. Studies have found characteristic signal changes when falls occur, distinct from regular walking~\cite{mercuri2012sfcw,erol2016wideband,peng2016fmcw}. 

The integration of textural-based feature extraction methods, wideband radars, and SVM have been shown to optimize false alarm problems in radar fall detection systems, achieving 95.0\% accuracy~\cite{erol2016wideband}. Utilizing Range-Doppler radars and deep learning, researchers have achieved a success rate of 97.1\% in minimizing false alarms~\cite{jokanovic2017fall}. Notably, this is the first paper to use Radar Cross-Section (RCS) information for fall detection~\cite{peng2016fmcw}. 

Researchers have shown that data fusion of two Ultrawide Band (UWB) radars using different fusion architectures improves performance and reduces false alarms, with Multi-Sensor feature-level fusion yielding the best results~\cite{erol2017range}. Other methods include using a TeraHertz (THz) FMCW radar to extract features and analyze with different classifiers~\cite{he2019human}, and using deformable deep CNN with a 1-class contrastive loss function achieving an accuracy of 99.5\%~\cite{shankar2019radar}. Lastly, the use of RadarNet, a structure of CNN followed by Inception modules, has shown an accuracy rate of 98.0\% in target classification for people or dogs~\cite{bhattacharya2020deep}. 

In conclusion, the application of Range-Doppler in fall detection has made significant progress, utilizing diverse radar technologies and advanced machine learning techniques. Integration of machine learning methods has improved accuracy and reliability, while innovative approaches like RCS information and sensor data fusion have reduced false alarms. Continued efforts are needed to enhance system robustness and adaptability. Please refer to Table~\ref{tab6} for a summary of fall detection methods with Range-Doppler.

\begin{table*}[!t]
\renewcommand{\arraystretch}{1.3}
\caption{Fall Detection with Range-Doppler}
\label{tab6}
\centering
\scriptsize
\definecolor{headercolor}{rgb}{0.88,0.88,0.88} 
\begin{tabularx}{1.0\textwidth}{|p{3cm}|p{1.5cm}|p{1.5cm}|p{1cm}|p{4cm}|X|}
\hline
\rowcolor{headercolor} \textbf{Paper} & \textbf{Radar} & \textbf{Classifier} & \textbf{Subjects} & \textbf{Motions} & \textbf{Performance}\\
\hline
Mercuri, M., et al.~\cite{mercuri2012sfcw} & SFCW & - & 1 & 3 activities & -\\
\hline
Erol, Baris, et al.~\cite{erol2016wideband}  & UWB & SVM & 4 & 4 activities, 106 samples & Prec=97.7\%, Rec=94.7\%, F1=96.2\%\\
\hline
Peng, Zhengyu, et al.~\cite{peng2016fmcw} & FMCW & - & 1 & 3 activities & -\\
\hline
Erol, Baris, et al.~\cite{erol2016fall} & UWB & SVM & 4 & 4 activities, 106 samples & Prec=99.6\%, Rec=99.6\%, F1=100.0\%\\
\hline
Wang, Haofei, et al.~\cite{wang2016phase}  & SFCW & - & 1 & 2 activities & -\\
\hline
Erol, Baris, et al.~\cite{erol2016range}  & UWB & SVM & 4 & 25 falls, 47 non-falls & Prec=100.0\%, Rec=95.0\%, F1=97.5\%\\
\hline
Erol, Baris, et al.~\cite{erol2017range} & UWB & kNN, SVM & 3 & 4 activities & Prec=97.0\%, Rec=95.0\%, F1=95.9\%\\
\hline
Jokanovic, Branka, et al.~\cite{jokanovic2017fall}  & FMCW & AE & 3 & 4 activities, 117 falls and 291 non-falls & Prec=88.2\%, Rec=78.9\%, F1=83.4\%\\
\hline
Erol, Baris, et al.~\cite{erol2018radar}  & UWB & kNN & 6 & 4 activities & Prec=99.1\%, Rec=96.6\%, F1=97.9\%\\
\hline
Ding, Chuanwei, et al.~\cite{ding2019fall}  & FMCW & kNN & 3 & 6 activities & Acc=95.5\%\\
\hline
Ding, Chuanwei, et al.~\cite{ding2019continuous} & FMCW & kNN & 8 & 6 activities & Acc=91.9\%\\
\hline
He, Mi, et al.~\cite{he2019human}  & FMCW & BPNN & 10 & 300 falls, 300 non-falls & Prec=84.3\%, AUC=0.932\\
\hline
Erol, Baris, et al.~\cite{erol2019radar} & FMCW & PCA, LDA, BPNN & 14 & 5 activities & Acc=97.2\%\\
\hline
Shankar, Yogesh, et al. ~\cite{shankar2019radar} & FMCW & CNN & 8 & 302 falls, 942 non-falls & Acc=99.5\%\\
\hline
Bhattacharya, Abhijit, et al.~\cite{bhattacharya2020deep} & FMCW & CNN & 2 & 3 activities & -\\
\hline
Wang, Bo, et al.~\cite{wang2020millimetre} & FMCW & CNN & 11 & 4 types of falls, 3 types of non-falls & Rec=99.6\%, Acc=98.7\%\\
\hline
Ma, Liang, et al.~\cite{ma2020room} & UWB & CNN-LSTM & 5 & 6 activities & Rec=98.0\%, Acc=95.8\%\\
\hline
Liang, Tingxuan, et al.~\cite{liang2021posture}  & FMCW & LSTM & - & 100 falls, 300 non-falls & Prec=98.9\%, Rec=99.0\%, F1=99.0\%\\
\hline
Ding, Chuanwei, et al.~\cite{ding2021sparsity}  & FMCW  & SVM & 3 & 6 activities & Acc=95.0\%\\
\hline
\end{tabularx}
\end{table*}

\subsection{Fall Detection with Range-Doppler-Angles}

FMCW radio's use of dual antenna arrays permits spatial separability, enabling differentiation of reflections at varied elevations.

\subsubsection{Range-Angles Heatmaps}

Aryokee, proposed by Tian et al.\cite{tian2018rf}, combines FMCW radar and spatio-temporal CNNs to distinguish fall and stand-up actions, delivering a recall of 94.0\% and a precision of 92.0\%. Sun et al.\cite{sun2019privacy} achieved superior performance using LSTM over 3-D CNN for processing range-angle heatmaps.

Ding et al.\cite{ding2021fall} used mmWave FMCW radar and kNN for fall classification in a 3D coordinate system. Su et al.\cite{su2022hybrid} employed a hybrid radar and AoA estimation to calculate biometrics and detect falls, regardless of the fall direction.

Yao et al.\cite{yao2022fall} utilized three neural networks for feature extraction from FMCW radar-generated maps, achieving a recall of 98.3\% and precision of 97.5\%. These approaches highlight the utility of range-angle data in radar-based fall detection. Refer to Table~\ref{tab7} for a summary.

\begin{table*}[!t]
\renewcommand{\arraystretch}{1.3}
\caption{Fall Detection with Range-Angle Heatmaps}
\label{tab7}
\centering
\scriptsize
\definecolor{headercolor}{rgb}{0.88,0.88,0.88} 
\begin{tabularx}{1.0\textwidth}{|p{3cm}|p{1.5cm}|p{1.5cm}|p{1cm}|p{4cm}|X|}
\hline
\rowcolor{headercolor} \textbf{Paper} & \textbf{Radar} & \textbf{Classifier} & \textbf{Subjects} & \textbf{Motions} & \textbf{Performance}\\
\hline
Tian, Yonglong, et al.~\cite{tian2018rf}  & FMCW & CNN & 140 & 40 activities & Prec=91.9\%, Rec=93.8\%, F1=92.9\%\\
\hline
Sun, Yangfan, et al.~\cite{sun2019privacy}  & FMCW & LSTM & 1 & 7 activities & Prec=100.0\%, Rec=93.6\%, F1=96.7\%\\
\hline
Ding, Congzhang, et al.~\cite{ding2021fall}  & FMCW & kNN & 5 & 150 falls, 150 non-falls & Acc=91.2\%\\
\hline
Su, Wei-Chih, et al.~\cite{su2022hybrid}  & CW & - & 1 & 3 activities & -\\
\hline
Yao, Yicheng, et al.~\cite{yao2022fall}  & FMCW & CNN & 21 & 12 types of falls, 52 types of non-falls & Prec=97.5\%, Rec=98.3\%, F1=97.7\%\\
\hline

\end{tabularx}
\end{table*}

\subsubsection{3D Point-Cloud}

Jin et al.\cite{jin2020mmfall} introduced an unsupervised fall detection method, mmFall, that uses a variational RNN Autoencoder (VRAE) for point cloud analysis, achieving a 98.0\% fall detection rate. Liang et al.\cite{liang2022fall} leveraged radar point clouds for human pose identification and 4G technology for data visualization on a cloud platform. These works underscore the potential of radar point clouds in accurate fall detection. Refer to Table~\ref{tab8} for a summary.

\begin{table*}[!t]
\renewcommand{\arraystretch}{1.3}
\caption{Fall Detection with 3D Point-Cloud}
\label{tab8}
\centering
\scriptsize
\definecolor{headercolor}{rgb}{0.88,0.88,0.88} 
\begin{tabularx}{1.0\textwidth}{|p{3cm}|p{1.5cm}|p{1.5cm}|p{1cm}|p{4cm}|X|}
\hline
\rowcolor{headercolor} \textbf{Paper} & \textbf{Radar} & \textbf{Classifier} & \textbf{Subjects} & \textbf{Motions} & \textbf{Performance}\\
\hline
Jin, Feng, et al.~\cite{jin2020mmfall}  & FMCW & VAE & 2 & 50 falls, 200 non-falls & Prec=96.1\%, Rec=98.0\%, F1=97.0\%\\
\hline
Liang, Jiye, et al.~\cite{liang2022fall}  & FMCW & BPNN & - & 2 activities & Prec=96.0\%, Rec=94.1\%, F1=95.2\%\\
\hline

\end{tabularx}
\end{table*}

FMCW radar, combined with machine learning techniques, has shown significant potential for accurate fall detection. Key methods include range-angle heatmaps and 3D point-cloud data, which effectively discern falls. Performance is further improved through algorithms like spatio-temporal CNNs. Additionally, hybrid technologies and unsupervised learning methods present innovative opportunities for this field. As research continues, these techniques are expected to advance, leading to more reliable fall detection solutions.

\section{Discussion}

Micro-Doppler methodologies have become prominent in fall detection research, largely because they easily extract signatures from radar Analog-to-digital conversion (ADC) raw data, present this data as images suitable for image processing techniques, and benefit from the early development of Doppler radar technology. Moreover, radar's most significant strength is the measurement of velocity. The integration of range information in 2016 was a pivotal development, facilitating superior target distinction and noise reduction capabilities. This breakthrough revitalized interest in the field and prompted further research efforts. Afterward, the MIMO technology and the advent of PCB-integrated antennas have significantly enhanced radar's capabilities, particularly in achieving three-dimensional spatial resolution. This has led to an upsurge in interest in Range-Doppler-Angles radar since 2018. Researchers are exploiting range-angle data, either through preserving original information via range-angle maps or using advanced 3D point cloud data for efficient real-time detection. 

While radar-based methods for fall detection offer promising outcomes, it is crucial to acknowledge their challenges and limitations: (i) Environmental Factors: Radar sensors can be affected by environmental factors, such as interference from other electronic devices, and metal materials. (ii) Calibration and Setup: Proper calibration is crucial for the effectiveness of the radar system. Changes in room layout or the introduction of new large objects can affect the system's performance. (iii) Hardware Limitations: The resolution of radar sensors can be affected by the hardware's quality and capability, and is relatively low. Higher-resolution systems often come at a higher cost, posing challenges in large-scale implementations. (iv) Real-world Applications: Despite experimental successes, transitioning these advancements to reliable real-world applications is an ongoing challenge. The variance observed between laboratory and real-world outcomes accentuates the need for bridging this gap.

Potential strategies for enhancing the efficacy of fall detection systems involve a multi-faceted approach tailored to specific application scenarios. It is crucial to select radar sensors that best match the desired performance criteria, such as optimal working distance, frame rate, and FoV. Depending on the room's layout and purpose, strategically placing sensors in locations that can easily capture fall events is essential. Defining the detection range and learning from known interference sources to apply area masking can further optimize detection. Moreover, customizing user parameters and establishing methods for device recalibration are valuable steps toward precision. Exploring techniques like lifelong/continual learning can help the system adapt over time, refining its performance based on continuous data ingestion. Periodically updating the system based on user feedback ensures that it remains responsive to real-world challenges and maintains user trust. Additionally, the fusion of data from multiple sensors offers a broader perspective, enhancing the system's accuracy and robustness in diverse conditions.

While implementing these strategies promises a more reliable and efficient fall detection system, the foundation of such advancements lies in robust research methodologies. This involves rigorous data collection protocols, consistent statistical analyses, and emphasizing results reproducibility. Equally important is the disclosure of specific radar parameters, including frequency range, power levels, and antenna specifications, which can greatly impact results. Promoting the sharing of datasets not only enables peer verification but also facilitates cumulative scientific understanding. Adherence to standardized assessment benchmarks, such as precision, recall, and F1 score, provides a common ground for evaluating the effectiveness of different approaches. Ideally, data collection should mirror real-world scenarios closely, thereby bolstering the real-life applicability of research insights. The development and availability of a holistic, large-scale radar fall detection dataset could serve as a cornerstone for the field, potentially attracting scholars from interdisciplinary backgrounds to contribute their expertise.

\section{Conclusion}
In this survey, we navigated the multifaceted domain of fall detection, initiating our exploration with an understanding of falls and progressing toward an analysis of technologies in the field. These form the foundation of today's radar-based fall detection methods, as evidenced by our extensive review of 74 pivotal studies that showcase the technological evolution, categorized into Micro-Doppler, Range-Doppler, and Range-Doppler-Angles detection techniques. We hope that this review offers a sense of the history, development, and potential future of radar for fall detection.
\section*{Acknowledgment}
We sincerely thank the authors we referenced and greatly value the constructive feedback from reviewers. We're also grateful for the financial support from the National Institutes of Health that enabled this research.
\bibliographystyle{IEEEtran}
\bibliography{ref}

\end{document}